\documentclass[journal]{IEEEtran}
\usepackage{amsmath}
\usepackage{lineno}
\usepackage{soul}
\usepackage{graphicx}
\usepackage{blindtext}
\usepackage{cite}
\usepackage{url}

\usepackage[english]{babel}
\usepackage[utf8]{inputenc}
\usepackage{fancyhdr}

\usepackage{booktabs}
\usepackage{multirow}
\usepackage[table,xcdraw]{xcolor}
\usepackage{placeins}
\usepackage[inline]{enumitem}
\usepackage[vlined,linesnumbered,ruled,resetcount]{algorithm2e}

\usepackage{bm}
\usepackage{tikz}
\usetikzlibrary{positioning,spy}
\usepackage{breqn}

\usepackage{float}
\usepackage[export]{adjustbox}
\usepackage[caption = false]{subfig}

\makeatletter
\newcommand{\clearsubcaptcounter}{\setcounter{sub\@captype}{0}}
\makeatother

\usepackage{etoolbox, siunitx}

\usepackage{amssymb}
\usepackage{pifont}
\newcommand{\cmark}{\ding{51}}%
\newcommand{\xmark}{\ding{55}}%

\usepackage{tabularx}
\usepackage{array}
\usepackage{rotating}

\newcolumntype{Y}{>{\centering\arraybackslash}X}

\ifCLASSINFOpdf
\else
\fi
\usepackage{fancyhdr}
\usepackage{lipsum}

\fancypagestyle{plain}{
  \fancyhf{}
  \fancyhead[C]{Conference on \LaTeX}     
  \fancyfoot[L]{This is a notice}

}
\usepackage{tikz}
\usetikzlibrary{calc}
\usepackage{atbegshi}

\begin{document}

%
\title{Self-Gated Memory Recurrent Network for Efficient Scalable HDR Deghosting}
%
%
%

\author{K~Ram~Prabhakar*,
        Susmit~Agrawal*,
        and~R~Venkatesh~Babu,~\IEEEmembership{Senior~Member,~IEEE}
\thanks{* equal contribution. \\ K.~R.~Prabhakar, S.~Agrawal and R.~V.~Babu are with the Department of Computational and Data Sciences, Indian Institute of Science, Bengaluru, KA, 560012, INDIA. e-mail: ramprabhakar@iisc.ac.in.

This paper has supplementary downloadable material available at \protect\url{http://ieeexplore.ieee.org}, provided by the author. The material includes additional results. This material is 15MB in size.
}
}

\urlstyle{tt}
\maketitle

\AtBeginShipout{\AtBeginShipoutAddToBox
{
    \begin{tikzpicture}[remember picture, overlay]
    \node at ($(current page.north) + (0,-0.15in)$) {\scriptsize This article has been accepted for publication in IEEE Transactions on Computational Imaging. This is the author's version which has not been fully edited and content may change prior to final publication.};
    \node at ($(current page.north) + (0,-0.25in)$) {\scriptsize Citation information: DOI 10.1109/TCI.2021.3112920};
    \node at ($(current page.south) + (0,+0.1in)$) {\scriptsize © 2021 IEEE. Personal use is permitted, but republication/redistribution requires IEEE permission. See https://www.ieee.org/publications/rights/index.html for more information.};
    \end{tikzpicture}
}}

\begin{tikzpicture}[remember picture, overlay]
    \node at ($(current page.north) + (0,-0.15in)$) {\scriptsize This article has been accepted for publication in IEEE Transactions on Computational Imaging. This is the author's version which has not been fully edited and content may change prior to final publication.};
    \node at ($(current page.north) + (0,-0.25in)$) {\scriptsize Citation information: DOI 10.1109/TCI.2021.3112920};
    \node at ($(current page.south) + (0,+0.1in)$) {\scriptsize © 2021 IEEE. Personal use is permitted, but republication/redistribution requires IEEE permission. See https://www.ieee.org/publications/rights/index.html for more information.};
    \end{tikzpicture}

\begin{abstract}
We propose a novel recurrent network-based HDR deghosting method for fusing arbitrary length dynamic sequences. The proposed method uses convolutional and recurrent architectures to generate visually pleasing, ghosting-free HDR images. We introduce a new recurrent cell architecture, namely Self-Gated Memory (SGM) cell, that outperforms the standard LSTM cell while containing fewer parameters and having faster running times. In the SGM cell, the information flow through a gate is controlled by multiplying the gate's output by a function of itself. Additionally, we use two SGM cells in a bidirectional setting to improve output quality. The proposed approach achieves state-of-the-art performance compared to existing HDR deghosting methods quantitatively across three publicly available datasets while simultaneously achieving scalability to fuse variable length input sequence without necessitating re-training. Through extensive ablations, we demonstrate the importance of individual components in our proposed approach. The code is available at \protect\url{https://val.cds.iisc.ac.in/HDR/HDRRNN/index.html}.
\end{abstract}


%
\IEEEpeerreviewmaketitle

\section{Introduction}
    \label{sec:intro}
    \IEEEPARstart{S}{tandard} digital cameras have limited sensor capability that can capture only a part of the natural scene illumination. This results in the captured images being too bright or too dark in certain regions, causing loss of textural and structural details in those regions. Specialized camera sensors capable of capturing natural high dynamic ranges are prohibitively expensive for day-to-day use cases \cite{tocci2011versatile,zhao2015unbounded}. A cheaper and more practical substitute is to provide a software alternative to advanced hardware. One commonly used software solution is to capture multiple Low Dynamic Range (LDR) images, each with a different exposure value, and combine them to generate an image containing a High Dynamic Range (HDR) of illumination. The generated HDR image has enriched details and textures in both highlight and shadow regions. 

In the absence of camera or object motion across the captured LDR images, HDR generation is a simple process \cite{debevec2008recovering}. However, if the images contain relative object or camera motion, fusing them with a standard HDR merging algorithm will result in ghost-like artifacts in the final result. The process of obtaining ghost-free HDR images even in the presence of movement among captured images is known as \textit{HDR deghosting}. The HDR deghosting problem has a rich history spanning two decades, and several methods have been proposed in the literature to address it. 

Rejection based methods attempt to remove pixels affected by motion (or dynamic pixels) from largely static exposure stacks. These dynamic pixels are replaced by LDR content from one of the captured images \cite{grosch2006fast,wu2010robust,gallo2009artifact,min2009histogram,heo2010ghost,raman2011reconstruction}. While these methods offer good quality results for mostly static scenes, they suffer from low HDR content in moving regions. Alignment based methods transform the input stack to resemble a chosen reference image, enforcing similar structure across all inputs \cite{tomaszewska2007image,bogoni2000extending,zimmer2011freehand,ward2003fast,gallo2015locally}. The aligned images are then fused using HDR merging methods such as \cite{debevec2008recovering}. The drawback is that most alignment methods are time and compute-intensive and are often unable to remove all artifacts. Patch-based optimization methods synthesize an HDR output that resembles a chosen reference image in well-exposed regions of reference and borrow details from other LDR images in regions where reference is poorly exposed \cite{sen2012robust,hu2013hdr,ma2017robust}. Despite their better performance, they are complex algorithms that suffer from high computational costs.

Data-driven methods use learning algorithms, primarily Convolutional Neural Networks (CNN), to learn fusion as well as deghosting \cite{kalantari2017deep,wu2018deep,prabhakar2019fast,yan2019multi,prabhakar2020}. Once trained, they can generate high-quality HDR images faster than most other non-deep methods. A potential drawback, however, is that most existing learning-based methods today are not \textit{scalable} \cite{kalantari2017deep,wu2018deep,yan2019multi,prabhakar2020}. We refer to scalability as a trait of a method to fuse arbitrary length LDR image sequences during inference. Most existing CNN-based methods are trained to operate on three images, with the middle image being the default reference. Hence, they can only fuse three LDR images during inference, not more, and not less than three.

In practical use cases, some challenging natural scenes may contain a very high illumination range that requires more than three LDR images to cover completely. In such cases, most existing CNN based HDR deghosting methods cannot be used to merge arbitrary length LDR images. These methods need to be re-trained with a new number of input images. However, such a solution poses two challenges: 1) A new and separate model must be trained for each possible sequence length. Hence, it increases the memory and storage footprint on the device. 2) New datasets have to be captured for different sequence lengths. Additionally, it is tedious to capture ground truth HDR, as it involves capturing static and dynamic sequences of the same scene with controlled motion. This shows the importance and need of having a method than can handle arbitrary exposure brackets and varying input lengths without necessitating re-training. 

\begin{figure}
    \centering
    \includegraphics[width=\linewidth]{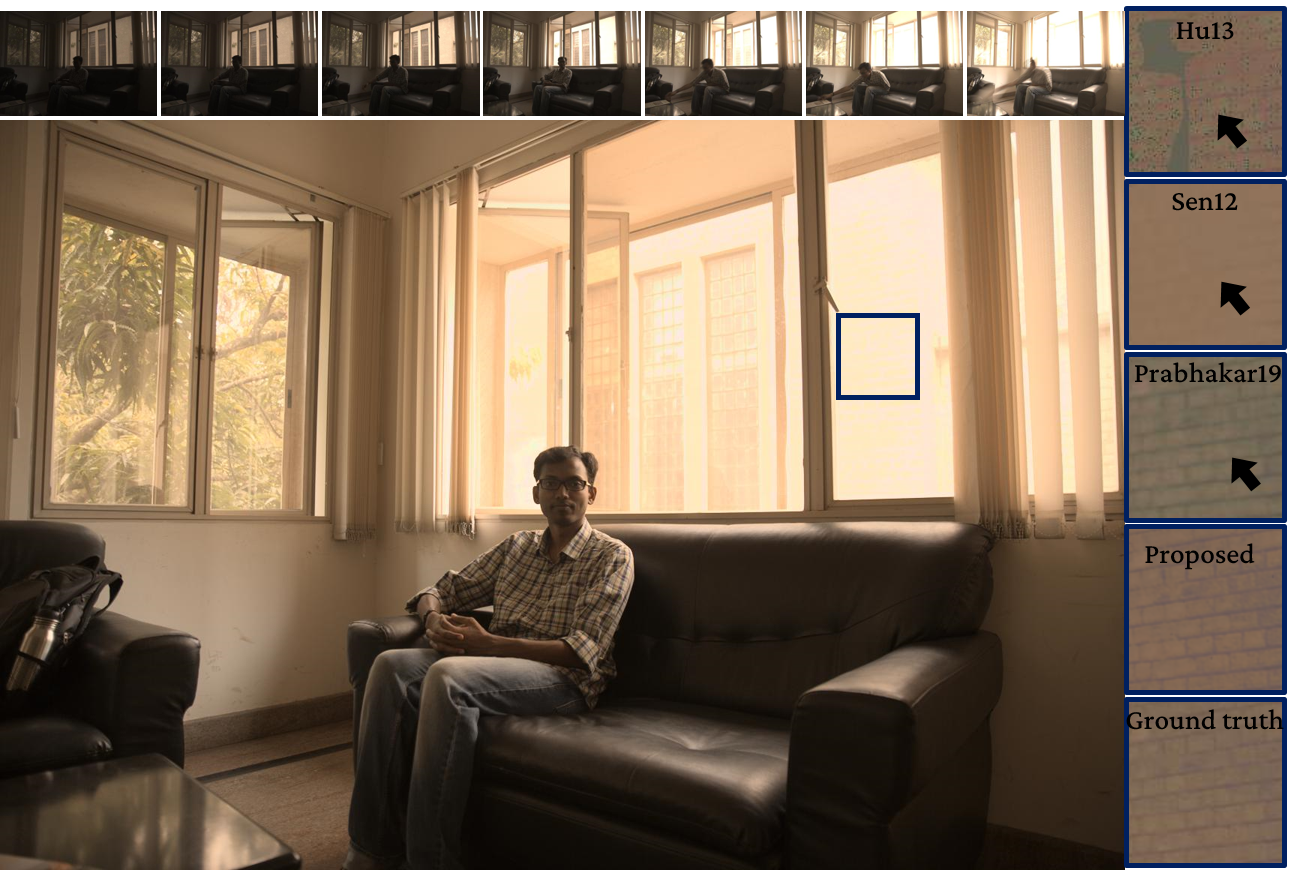}
    \caption{A qualitative example to highlight results by various scalable HDR deghosting methods on a sequence with seven dynamic LDR images (top row). Existing scalable HDR deghosting methods are either too slow or introduce visible artifacts (highlighted by black arrow). Our proposed method generates high-quality artifact-free HDR images with less computational cost.} 
    \label{fig:seq7_1}
\end{figure}

Among state-of-the-art non-deep methods, Sen \textit{et al.} \cite{sen2012robust} and Hu \textit{et al.} \cite{hu2013hdr} methods are scalable. However, these methods have significant computation overhead and generate images with artifacts (see Fig. \ref{fig:seq7_1}). The CNN-based method proposed by Prabhakar \textit{et al.} \cite{prabhakar2019fast} is capable of fusing a variable number of images, but at a considerable cost in terms of quality due to their internal aggregation strategy of features (see Fig. \ref{fig:iccp2}c and \ref{fig:seq7_2}). This indicates that there is still room for developing a better and efficient scalable deep HDR deghosting method. 

We address the problem as mentioned above by exploiting the internal memory capabilities of recurrent architectures. In our proposed approach, we introduce a novel recurrent cell called Self-Gated Memory (SGM) cell for efficient HDR deghosting. The SGM cell's internal vector representations make an ideal way to encode pixel color and intensity data while simultaneously filtering out structural information of non-reference images. Since the misaligned structures in non-reference images are the major cause of ghosting artifacts, removing them results in higher quality images. Also, we improve the accuracy of our model by using two SGM cells in a bidirectional setting. 
In summary, the main contributions of our work are as follows:
\begin{itemize}[noitemsep,topsep=0pt,itemsep=0pt]
  \item We propose a novel network that uses recurrent architectures for efficient and scalable HDR deghosting. Our network can fuse an arbitrary number of LDR images into a ghost-free HDR image without re-training. To the best of our knowledge, our model is the first RNN-based approach proposed for multi-shot HDR deghosting.

  \item We present a novel Self-Gated Memory recurrent cell that uses self-gating to control data flow. The proposed SGM cell outperforms existing state-of-the-art HDR deghosting methods and other standard recurrent architectures. 

  \item We perform extensive experiments to demonstrate the superiority of the proposed approach. Additionally, through rigorous ablation experiments, we justify the importance of each module in our proposed method.
\end{itemize}

The rest of the paper is organized as follows. Section \ref{sec:related} presents a brief review of existing HDR deghosting methods in the literature. In Section \ref{sec:prop}, we present our novel recurrent architecture for HDR deghosting and describe the methodology in detail. In Section \ref{sec:eval}, we compare the proposed approach with existing state-of-the-art deghosting methods. In Section \ref{sec:discussion}, we present the discussion on several ablations and running time comparison. Finally, we conclude the paper in Section \ref{sec:conc}.  
\section{Related Works}
    \label{sec:related}
    Naively fusing images with camera and object motion will result in visible ghosting artifacts. Several methods have been proposed in the past to avoid such ghosting artifacts\footnote{Please refer to \cite{sen2016practical,sen2018overview,tursun2015state} for more detailed literature review.}.

\textit{Alignment methods}: This class of algorithms aligns all input images with camera motion to a chosen reference image. Thus, the generated image sequence is mostly static and can be fused using standard HDR fusion algorithms with minimal ghosting artifacts. These methods are not primarily deghosting approaches suitable to handle object motion, hence they are used as a pre-processing step for such algorithms to eliminate global camera or scene motion. The methods within this category mostly differ by the choice of alignment strategy such as frequency domain matching and cross correlation \cite{cerman2006exposure,rad2007multidimensional,yao2011robust}, elastic registration \cite{im2011geometrical,im2011improved}, SIFT \cite{tomaszewska2007image}, CIFT \cite{gevrekci2007geometric}, neighbourhood similarity \cite{akyuz2011photographically}, etc. The early work by Mann \textit{et al.} \cite{mann2002painting} proposed to jointly estimate CRF and register images using global homography. Candocia \textit{et al.} \cite{candocia2003simultaneous} register inputs by optimizing to reduce pixel intensity variance. An alignment method proposed by Tomaszewska and Mantiuk \cite{tomaszewska2007image} use SIFT to perform a key-point search in consecutive images of the input stack, and then use RANSAC to improve the matching. The images are warped using a homography computed from the matched key-points. 

Ward \cite{ward2003fast} proposed an algorithm that runs in linear time and uses grayscale images to create image pyramids. Bogoni \cite{bogoni2000extending} perform both global and local alignment in two steps using affine transforms and optical flow correction. Zimmer \textit{et al.} \cite{zimmer2011freehand} propose a method involving the optimization of a two-term energy function, containing a data term to promote alignment to reference and a regularizer term to ensure a smooth flow in poorly exposed regions. This optimization is applied after aligning images with optical flow. A fast alignment method introduced by Gallo \textit{et al.} \cite{gallo2015locally} computes sparse correspondences between an input and the reference image and propagates the sparse flow in an edge-aware fashion to approximate the dense flow map. Kang \textit{et al.} \cite{kang2003high} globally align frames in a video sequence and then use gradient-based optical flow to correct local regions to generate HDR video. 

\textit{Rejection based methods}: This class of algorithms works well with mostly static image stacks \cite{srikantha2012svd,heo2010ghost,sung2013feature}. The static pixels that are void of motion are fused using traditional static HDR merging techniques such as \cite{debevec2008recovering}. The algorithms in this category can be further divided into two subgroups, depending on how they identify and handle dynamic pixels. The first subclass of these algorithms replaces moving regions with content from the chosen reference image. Grosch \cite{grosch2006fast} generate an error map based on thresholded color differences between pixels. The generated error map represents moving regions, which can then be filled from the reference image. An \textit{et al.} \cite{an2011multi} use zero-mean normalized cross-correlation to obtain dynamic pixels between input images. Wu \textit{et al.} \cite{wu2010robust} find pixels that violate brightness consistency to detect dynamic regions. Gallo \textit{et al.} \cite{gallo2009artifact} threshold the difference in the logarithmic domain instead of directly using pixel values to identify regions affected by motion. 

Lee \textit{et al.} \cite{lee2011improved} detect moving pixels by thresholding the difference of rank normalized input images. Li \textit{et al.} \cite{li2010movement} compute bi-directional pixel similarity metric between input and reference to identify motion region. Min \textit{et al.} \cite{min2009histogram} identify moving regions by performing multi-level thresholding of intensity histograms and use these regions to construct a radiance map. Raman \textit{et al.} \cite{raman2011reconstruction} group superpixels to identify dynamic regions. The result is used to create a largely static version of the input stack. Heo \textit{et al.} \cite{heo2010ghost} locate the ghosted pixels by thresholding the joint probability computed between the reference and the other source images. Lin \textit{et al.} \cite{lin2009high} threshold the difference between normalized input images to find motion regions. 

The second subclass of rejection based methods contain algorithms that work without selecting a reference image from the inputs \cite{zhang2011gradient,silk2012fast,zhang2010gradient}. They ignore images affected by the motion for the dynamic pixels and only use the rest of the stack images. The final result contains static regions of the whole scene. Khan \textit{et al.} \cite{khan2006ghost} use a kernel density estimator function to compute pixel weights and iteratively optimize the function. Their approach does not need explicit object detection or motion estimation. Pedone \textit{et al.} \cite{pedone2008constrain} improve deghosting performance of \cite{khan2006ghost} with morphological operations on the estimated motion bitmaps. Pece \textit{et al.} \cite{pece2010bitmap} detect clusters of pixels affected by motion using binary operations and pick the least saturated clusters. Eden \textit{et al.} \cite{eden2006seamless} use a graph-cut method to address object motion within the input stack. 

Granados \textit{et al.} \cite{granados2010optimal} iteratively reconstruct the HDR image by estimating a mean radiance map with minimum variance. An earlier work by Reinhard \textit{et al.} \cite{reinhard2005dynamic} utilizes normalized variance to identify motion affected pixels. Building on top of \cite{reinhard2005dynamic}'s approach, Jacobs \textit{et al.} \cite{jacobs2008automatic} proposed to use an entropy measure instead of variance. Sidibe \textit{et al.} \cite{sidibe2009ghost} increase the exposure and locate dynamic pixels that does not proportionally increase its intensity value. Oh \textit{et al.} \cite{oh2014robust} proposed a rank minimization approach to generate final HDR output. In their approach, authors calculate rank-1 matrix of stacked input images. The residual noise of the estimate is used to locate the moving dynamic pixels. 

\textit{Non-rigid registration methods}: The following set of methods use optical flow to handle both global camera motion as well as local object motion. These algorithms warp moving regions in images of the input stack, such that these regions align with a chosen reference image. The final HDR result is obtained by merging the aligned static sequence. Ferradans \textit{et al.} \cite{ferradans2012generation} use GMM to model the difference between optical flow warped images. The pixels that do not fit to estimated modes within pre-defined variance receives less weight in the fusion process. Jinno \textit{et al.} \cite{jinno2011multiple} jointly estimate flow, occlusion and saturation maps by minimizing an energy function, which are later used to fuse the images. Hafner \textit{et al.} \cite{hafner2014simultaneous} jointly estimate fused HDR and displacement fields in a optimization framework enforced with spatial smoothness constraints. This category methods suffer from erroneous dense correspondence or optical flow due to complex non-rigid motions and occlusion. 

\textit{Patch based optimization methods}: This class of methods generates a final HDR result similar to the reference image in regions where the reference image is properly exposed \cite{menzel2007freehand,park2011motion}. Sen \textit{et al.} \cite{sen2012robust} propose a method to optimize both structure and content of the HDR image. They achieve it by parameterizing these quantities in an image synthesis equation and optimize for the same. Hu \textit{et al.} \cite{hu2013hdr} generate latent images from the input stack that are similar in structure but vary in an exposure. While this class of methods addresses the shortcomings of rejection based methods and registration methods, they suffer from high computational complexity. 

\textit{Data driven techniques}: These approaches make use of Convolutional Neural Networks trained on data that captures the essence of the HDR deghosting problem. Fusion and Deghosting rules are approximated by learning from examples. Kalantari \textit{et al.} \cite{kalantari2017deep} use an input stack that has been aligned using optical flow and train the model to correct warping artifacts. Wu \textit{et al.} \cite{wu2018deep} directly use neural networks to learn both alignment and fusion, thus attempting to remove the overhead of optical flow. Yan \textit{et al.} \cite{yan2019attention} propose attention mechanisms to focus only on the relevant information from the input stack. 

Yan \textit{et al.} \cite{yan2020deep} use a network with a non-local module to identify matching neighbor features to fill in ill-exposed regions of the reference image. Prabhakar \textit{et al.} \cite{prabhakar2019fast} aggregate features derived by shared CNN modules to create a scalable architecture that can fuse an arbitrary number of images without retraining. Prabhakar \textit{et al.} \cite{prabhakar2020} propose a method to minimize the consumption of memory and enable the fusion of high-resolution images. They perform alignment on low-resolution inputs and upsample intermediate features while generating the final HDR using a Bilateral Grid Upsampler \cite{gharbi2017deep}. In spite of their better performance, most of the existing deep-learning methods \cite{kalantari2017deep,wu2018deep,yan2019attention,yan2019multi,yan2020deep,prabhakar2020} are not scalable to fuse arbitrary number of images without re-training. 

\textit{Recurrent Neural Networks} have been extensively explored for processing temporal data such as speech \cite{zhao2017recurrent,pascual2016multi,gelly2017optimization}, natural language processing \cite{mikolov2012context,sundermeyer2014translation,ghosh2016contextual}, video captioning \cite{pei2019memory,yu2016video}, video segmentation \cite{hu2018maskrnn,valipour2017recurrent}, low-level video processing \cite{godard2018deep,chen2016deep} and in so many other applications. The closest matching related application to HDR deghosting that utilizes recurrent architecture is burst denoising approach by Godard \textit{et al.} \cite{godard2018deep} and the video denoising method by Chen \textit{et al.} \cite{chen2016deep}. To the best of our knowledge, there has not been any previous attempt to use recurrent networks for HDR deghosting. 

\section{Proposed method}
    \label{sec:prop}
    \subsection{Method overview}
\begin{figure*}[ht]
    \centering
      \includegraphics[width=\linewidth,clip, trim=2.425cm 0.2cm 0.85cm 0.4cm]{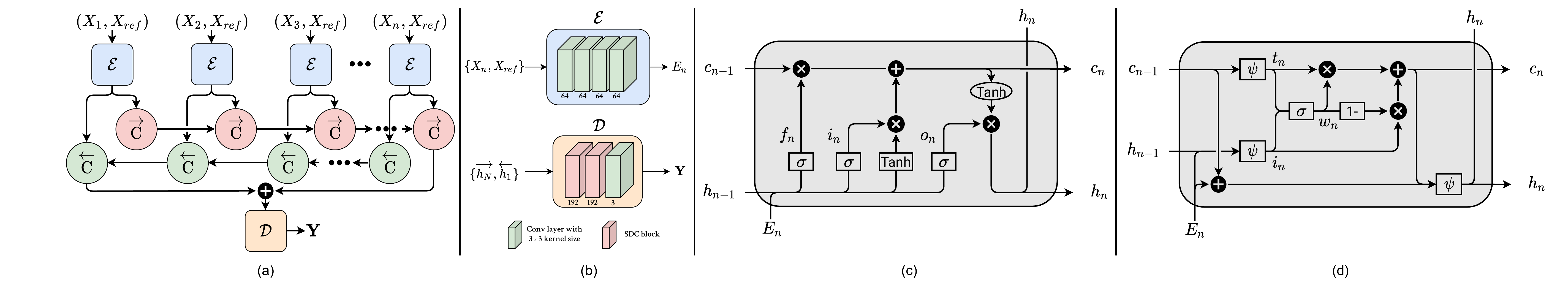}%
      \caption{(a) The architecture of the proposed approach. The current input data $X_n$ concatenated with the reference data $X_{ref}$ is passed as input to the encoder ($\mathcal{E}$). The encoded features are fed to the proposed bi-directional SGM recurrent cells ($\protect\overrightarrow{\text{C}}$ and $\protect\overleftarrow{\text{C}}$). The final output of the recurrent cells in both directions are concatenated and passed to the decoder ($\mathcal{D}$) to generate fused HDR ($\mathbf{Y}$). (b) Architecture of encoder ($\mathcal{E}$) and decoder ($\mathcal{D}$). The number below each block denotes the number of output channels of the corresponding block. Architecture of (c) standard LSTM cell, and (d) proposed Self-Gated Memory (SGM) recurrent cell.}
      \label{fig:prop_arch}
    \end{figure*}
Given a list of $N$ Low Dynamic Range (LDR) images $S = \{I_1, I_2,\cdots, I_N\}$, the goal of our approach is to merge them into a single HDR image $\mathbf{Y}$ without any ghosting artifacts. Assuming that the input sequence images are camera motion aligned, the challenge is to combine these images while accounting for object motion. Our proposed method is a reference-based method, where the final result will be structurally similar to one of the input images chosen as reference ($I_{ref}$). In particular, for static regions (regions without any object motion), the result contains HDR content fused from all input images. For dynamic regions (regions affected by motion), the result will have the same structure as the chosen reference image but with HDR content added from other images. Typically, the image with the least saturated pixels is chosen as a reference.

    
\textbf{Motivation}: Most deep learning-based approaches (\nolinebreak\hspace{1sp}\cite{kalantari2017deep,wu2018deep,yan2019attention,prabhakar2020}) are trained to fuse fixed number of LDR images, usually 3 images. These methods require re-training with the corresponding number of images for fusing a sequence with the different number of input images. Another shortcoming is the lack of a training dataset for different sequence lengths. It is very tedious to collect large labeled datasets for different sequence lengths, as it requires capturing both static and dynamic sequences of same scene. Prabhakar \textit{et al.}\cite{prabhakar2019fast} addressed scalability in their work by concatenating the mean and max of input features. However, their method still suffers from artifacts in challenging scenarios (shown in Fig. \ref{fig:seq7_1}).

Our proposed approach consists of a single neural network that uses convolutional and recurrent architectures. As shown in Fig. \ref{fig:prop_arch}a, the overall network can be logically divided into three sub-modules: an encoder, two recurrent cells in a bidirectional setting, and a decoder.

\textbf{Encoder ($\mathcal{E}$)}: As shown in Fig. \ref{fig:prop_arch}b, the encoder is a block of 3$\times$3 convolutional layers. It acts as a basic feature extractor and processes the input images for fusion by the recurrent module. Since the number of input images can vary, we use a single encoder to process all $N$ images in the input sequence. For the rest of the paper, let $I_n$ represent the $n^{th}$ image in the input sequence. The gamma-corrected HDR version ($H_n$) of $I_n$ is obtained by $H_n=I_n^{2.2}/t_n$, where $t_n$ is the exposure time of $I_n$. Let $X_n$ be the concatenation of $I_n$ and $H_n$. $\mathcal{E}$ then takes concatenated $X_n$ and $X_{ref}$ as input,
\begin{align}
    X_n &= I_n \oplus H_n \\
    X_{ref} &= I_{ref} \oplus H_{ref} \\
    E_n &= \mathcal{E}\big(\{X_n, X_{ref}\})
    \label{eqn:eqn1}
\end{align}
where, $\oplus$ denotes the feature concatenation operation along the channel axis.

\textbf{Recurrent Module}: This is the core module in the proposed architecture. The recurrent module takes as input the list of encoded images ${E_1, E_2, ..., E_n}$ and iteratively fuses them into output vectors ${h_1, h_2, ..., h_N}$, such that an output vector $h_n$ at the $n^{th}$ timestep contains information from all input vectors $E_1$ to $E_n$. The final output $h_N$ thus contains fused information from the entire input sequence.

The recurrent module may contain a single recurrent cell in a unidirectional setting, or it may use two cells in a bidirectional arrangement. In the bidirectional setting, one of the cells processes the input sequence starting from $E_1$ to $E_N$, while the second cell processes inputs starting from $E_N$ to $E_1$. The outputs for the cells are $\overrightarrow{h_N}$ and $\overleftarrow{h_1}$ respectively, with the arrows indicating the direction in which the input sequence is processed. The module's output is then given by concatenating the two vectors $\overrightarrow{h_N}$ and $\overleftarrow{h_1}$.

\begin{figure*}[t]
        \centering
        \includegraphics[width=\linewidth]{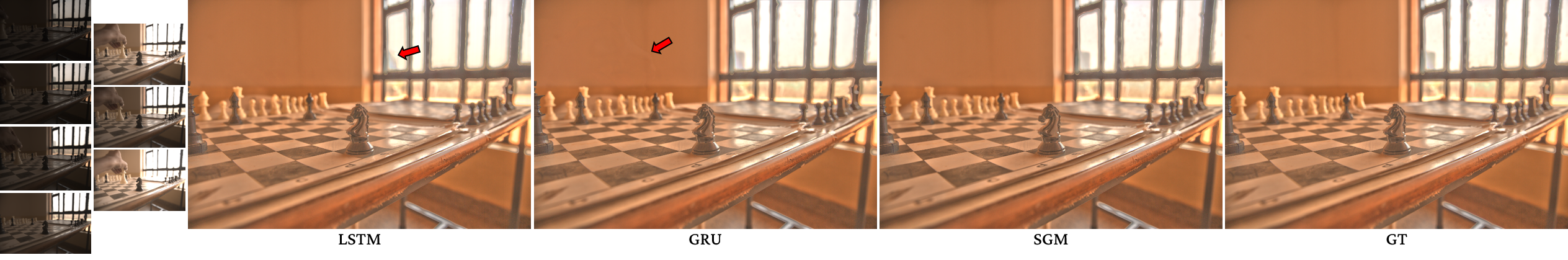}
        \caption{Qualitative comparison between proposed SGM cell against baseline recurrent cell types: LSTM and GRU.}
        \label{fig:rebuttal_fig5}
    \end{figure*}
In the forward direction, $h_0$ and $c_0$ are zero vectors of the same height and width as each input image, named $\overrightarrow{h_{init}}$ and $\overrightarrow{c_{init}}$. Then, in the reverse direction, $h_{N+1}$ and $c_{N+1}$ are $\overleftarrow{h_{init}}$ and $\overleftarrow{c_{init}}$ respectively. We use cell $\overrightarrow{\text{C}}$ in forward and a separate cell $\overleftarrow{\text{C}}$ in reverse direction for images $i=(1,\cdots,N)$,
\begin{align}
(\overrightarrow{h_i}, \overrightarrow{c_i}) &= \overrightarrow{\text{C}}\big(E_i,\overrightarrow{h_{i-1}},\overrightarrow{c_{i-1}} \big),\\
(\overleftarrow{h_{N-i+1}}, \overleftarrow{c_{N-i+1}}) &= \overleftarrow{\text{C}}\big(E_{N-i+1},\overleftarrow{h_{N-i+2}},\overleftarrow{c_{N-i+2}} \big)
\end{align}
For $N$=3, the forward cell unrolls as,
\begin{align}
    (\overrightarrow{h_1}, \overrightarrow{c_1}) &= \overrightarrow{\text{C}}\big(E_1,\overrightarrow{h_{init}},\overrightarrow{c_{init}} \big),\\
    (\overrightarrow{h_2}, \overrightarrow{c_2}) &= \overrightarrow{\text{C}}\big(E_2,\overrightarrow{h_1},\overrightarrow{c_1} \big),\\
    (\overrightarrow{h_3}, \overrightarrow{c_3}) &= \overrightarrow{\text{C}}\big(E_3,\overrightarrow{h_2},\overrightarrow{c_2} \big)
\end{align}
Whereas the reverse cell unrolls as,
\begin{align}
    (\overleftarrow{h_3}, \overleftarrow{c_3}) &= \overleftarrow{\text{C}}\big(E_3,\overleftarrow{h_{init}},\overleftarrow{c_{init}} \big),\\
    (\overleftarrow{h_2}, \overleftarrow{c_2}) &= \overleftarrow{\text{C}}\big(E_2,\overleftarrow{h_3},\overleftarrow{c_3} \big),\\
    (\overleftarrow{h_1}, \overleftarrow{c_1}) &= \overleftarrow{\text{C}}\big(E_1,\overleftarrow{h_2},\overleftarrow{c_2} \big)
\end{align}

\textbf{Feature decoder ($\mathcal{D}$)}: After processing the complete input sequence, the output features produced by the recurrent module are passed to a decoder block that generates the final HDR image. In bidirectional architecture, the concatenation of the final outputs from both recurrent cells serves as the input to this module (Fig. \ref{fig:prop_arch}b),
\begin{alignat}{2}
    \mathbf{Y} = \mathcal{D}(\overrightarrow{h_N} \oplus \overleftarrow{h_1})
    \label{eqn:eqn3}
\end{alignat}

In our implementation, we use two stacked SDC blocks \cite{schuster2019sdc} to capture details across different receptive fields, followed by a single convolutional layer to generate the final image.

\textbf{Loss}: The model produces an HDR image $\mathbf{Y}$ as its output. This image is tonemapped using the $\mu$-law tonemapping function defined as,
\begin{equation}
    T(\mathbf{Y}) = \dfrac{log(1+\mu \mathbf{Y})}{log(1+\mu)}
    \label{eqn:log_tm}
\end{equation}
where $\mu = 5000$.
The loss $\mathcal{L}$ between the ground truth HDR $\mathbf{\hat Y}$ and the predicted image $\mathbf{Y}$ is obtained by computing the $\ell_2$ loss between their tonemapped representations:
\begin{equation}
    \mathcal{L} = \ell_2(T(\mathbf{Y}), T(\mathbf{\hat{Y}}))
\end{equation}
\subsection{Self-Gated Memory (SGM) recurrent cell}
Before we present the details of the proposed SGM cell, we first discuss existing standard recurrent cells, their shortcomings and finally on how they are addressed in the SGM cell. We start with the three standard recurrent cell types, the standard RNN cell, the Long Short-Term Memory (LSTM), and the Gated Recurrent Unit (GRU) cells. Out of all three, LSTM performed better than other two cell types in terms of quantitative metrics (Table \ref{tab:ablation_arbitrary_length}). Qualitatively, images generated by LSTM has less ghosting artifacts compared to GRU (Fig. \ref{fig:rebuttal_fig5}). However, we identify three challenges in LSTM cell design that can be addressed to improve the performance specific to HDR deghosting task.

\begin{figure}[t]
        \centering
        \includegraphics[width=0.9\linewidth]{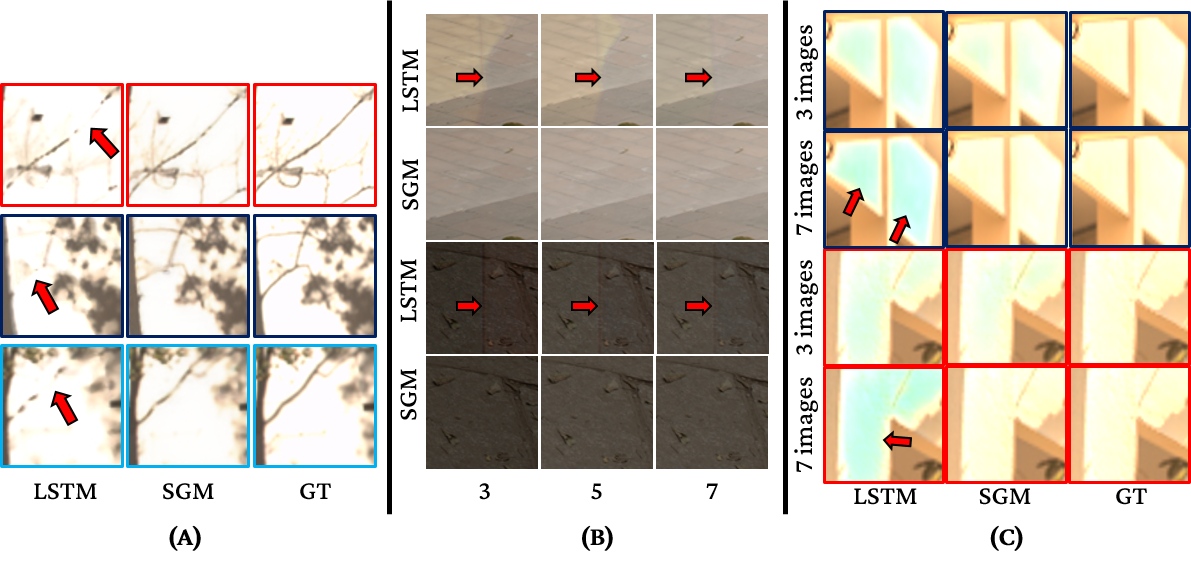}
        \caption{Qualitative comparison between LSTM and SGM cells.}
        \label{fig:lstm_compare}
    \end{figure}

The problems encountered with LSTM cell are discussed below, along with the corresponding change in the proposed SGM cell architecture to address the issue:
\begin{enumerate}[label=(\roman*),wide,itemindent=1em,noitemsep,topsep=0pt,itemsep=0pt]
    \item The forgetting of information from the internal state in both LSTM and SGM can be represented in the form: $c_n = \alpha \cdot c_{n-1} + \beta$. It is evident that $\alpha$ plays a large role in deciding what information is forgotten. In case of LSTM, $\alpha$ is obtained from the forget gate as shown below:
    \vspace*{-0.3cm}
    \begin{table}[h]
    \centering
    \begin{tabular}{l|l}
    \multicolumn{1}{c|}{LSTM} & \multicolumn{1}{c}{SGM} \\
    $\alpha = \sigma((h_{n-1} \oplus E_n) \odot k + b)$ & \begin{tabular}[c]{@{}l@{}}$\alpha = \sigma((t_n \oplus i_n) \odot k + b)$\\ $t_n = \psi(c_{n-1} \odot k + b)$\end{tabular}
    \end{tabular}
    \end{table}
    \vspace*{-\baselineskip}
    
    It is obvious through the equations that $c_{n-1}$ is not directly involved in deciding what information is to be ``forgotten" (Fig. \ref{fig:prop_arch}c). Thus, the important exposure information may be forgotten without much importance given to $c_{n-1}$. The forgotten information is then overwritten with information from the current image, resulting in a large contribution of final images to the output. In some cases, this manifests as saturated regions in the output, even though intermediate images may have the required detail to generate saturation-free results.
    
    To prevent this, we use $c_{n-1}$ in computation of the weight maps used for forgetting in the SGM cell. In the SGM cell, as shown in above table and in Fig. \ref{fig:prop_arch}d, $c_{n-1}$ contributes significantly in computation of the new state $c_n$. Visually, we see that SGM cell has retained structural information in locations where LSTM fails to, as shown in Fig. \ref{fig:lstm_compare}A. Quantitatively, cell architecture ablations types 6 and 7 in Table \ref{tab:sgm_cell_ablation} show that not using $c_{n-1}$ actively for computing the new state results in degraded performance.
    %
    %
    %
    \item LSTM performs minimal computation before deleting information from the long-term memory. The bulk of computation is performed in the input gate, and the result is unconditionally integrated with $c$. If any structural artifacts are not removed by the input gate, they will propagate through the rest of the sequence with high probability. These show up as warping artifacts in the final output of the LSTM cell, as shown in Fig. \ref{fig:lstm_compare}B.
        
    To address this issue in the SGM, we propose a set of architectural changes. First, we have individual layers for both $h$ and $c$, resulting in preliminary filtering from both long-term and short-term memory before state update. Next, we use both states to compute the weight map that ultimately updates $c$. It can be seen in Fig. \ref{fig:lstm_compare}B that this design helps to eliminate warping artifacts. The cell architecture ablations for these changes are Type 4, Type 6 and Type 7.
        
    \item Finally, LSTM maps high values generated internally as well as in its output to $[-1, 1]$ range, due to the Tanh activations. This means that if two pixel values from two different images are high, they will be mapped close together, losing relative intensity information between them. This is especially problematic if the number of images during evaluation is higher than the number of training images since they will probably have more over-saturated regions than what the LSTM has seen during training. In such cases, having additional images will not result in higher quality output, as shown in Fig. \ref{fig:lstm_compare}C. Both models were trained with only 3 images and tested on 3 and 7 length sequences. As highlighted by the arrows, an LSTM model trained with 3 images, generates over-saturation when tested with 7 images. In contrast, SGM cell can generate output without such artifacts.
    %
        
    SGM implicitly addresses this due to the Swish activation, which has an unbounded upper limit. Fig. \ref{fig:lstm_compare}C shows that SGM is able to extract information from long sequences, where LSTM saturates in quality.
\end{enumerate}

With these major considerations, we perform a series of cell architecture ablations to empirically get the optimal architecture for our application. The final cell contains the following structures:

\textit{Input gate}: This gate is used to extract relevant features from the input $E_n$ to the cell, and its previous output $h_{n-1}$. It uses a convolution block followed by self-gating:
\begin{alignat}{2}
    \overrightarrow{i_n} = \psi((\overrightarrow{h_{n-1}} \oplus E_n) \odot \overrightarrow{k_i} + \overrightarrow{b_i})
    \label{eqn:eqn15}
\end{alignat}
where $k_i$ is the kernel used to convolve the input, $b_i$ is the bias, $\oplus$ is the concatenation operation, and $\odot$ is the convolution operation.

\textit{Transform gate}: This gate is used to extract relevant features from the previous internal state $c_{n-1}$ of the cell. It uses a convolution block followed by self-gating:
\begin{alignat}{2}
    \overrightarrow{t_n} = \psi(\overrightarrow{c_{n-1}} \odot \overrightarrow{k_t} + \overrightarrow{b_t})
    \label{eqn:eqn16}
\end{alignat}
where $k_t$ is the kernel used to convolve the previous state, and $b_t$ is the bias.
    
\textit{Update gate}: The update gate is used to update the internal state $c_n$ of the cell, using the features extracted from the previous state $c_{n-1}$, previous output $h_{n-1}$ and the input $E_n$. It performs a weighted sum of the two features to derive the new state $c_n$. The weights are determined using a convolutional block having sigmoid activation:
\begin{align}
    \overrightarrow{w_n} &=\sigma((\overrightarrow{i_n} \oplus \overrightarrow{t_n}) \odot \overrightarrow{k_w} + \overrightarrow{b_w})\\
    \overrightarrow{c_n} &= \overrightarrow{w_n} \!\cdot\!\overrightarrow{t_n}\!+\![1\!-\!\overrightarrow{w_n}]\!\cdot\!\overrightarrow{i_n}
    \label{eqn:eqn17}
\end{align}
where $k_w$ is the kernel used to convolve the input, and $b_w$ is the bias.
\textit{Output Gate}: This gate produces the final output, $h_n$. It uses a convolution block followed by self-gating. The inputs to this gate are the old cell state ($c_{n-1}$), the new state ($c_n$) and the current input ($E_n$):
\begin{alignat}{2}
    \overrightarrow{h_n} = \psi((\overrightarrow{c_{n-1}} \oplus (\overrightarrow{c_n} + E_n)) \odot \overrightarrow{k_o} + \overrightarrow{b_o})
    \label{eqn:eqn19}
\end{alignat}
where $k_o$ is the kernel used to convolve the previous state, and $b_o$ is the bias.

We use two cells in a bidirectional arrangement in our implementation, each having 64 filters with 3$\times$3 kernel size for all its convolutional layers. The equations for the cell operating on the sequence $\mathcal{S}$ in the reverse direction are identical to the equations mentioned above, applied from $I_N$ to $I_1$:
\begin{align}
    \overleftarrow{i_{n}} &= \psi((\overleftarrow{h_{n+1}} \oplus E_n) \odot \overleftarrow{k_i} + \overleftarrow{b_i}) \\
    \overleftarrow{t_{n}} &= \psi(\overleftarrow{c_{n+1}} \odot \overleftarrow{k_t} + \overleftarrow{b_t}) \\
    \overleftarrow{w_{n}} &= \sigma((\overleftarrow{i_{n}} \oplus \overleftarrow{t_{n}}) \odot \overleftarrow{k_w} + \overleftarrow{b_w}) \\
    \overleftarrow{c_{n}} &= (\overleftarrow{w_{n}} * \overleftarrow{t_{n}}) + ((1 - \overleftarrow{w_{n}}) * \overleftarrow{i_{n}}) \\
    \overleftarrow{h_{n}} &= \psi((\overleftarrow{c_n} \oplus (\overleftarrow{c_{n+1}} + E_n)) \odot \overleftarrow{k_o} + \overleftarrow{b_o})
\end{align}
\begin{table}[t]
    \setlength{\tabcolsep}{3pt}
    \centering
    \caption{Comparison of various SGM cell architectures.}
    \label{tab:sgm_cell_ablation}
    \begin{tabular}{@{}llccc|ccc@{}}
    \toprule
    \multicolumn{2}{c}{Metrics ($\rightarrow$)} & \multicolumn{3}{c|}{PSNR-L} & \multicolumn{3}{c}{PSNR-T} \\ \midrule
    \multicolumn{2}{c}{\begin{tabular}[c]{@{}c@{}}Images ($\rightarrow$)\\ / Cell types ($\downarrow$)\end{tabular}} & 3 & 5 & 7 & 3 & 5 & 7 \\ \midrule
    &LSTM & 43.10 & 43.90 & 50.30 & 47.02 & 47.63 & 48.28\\ \midrule
    &Type1 & 43.21 & 43.69 & 49.25 & 46.61 & 47.37 & 47.06 \\
    &Type2 & 42.68 & 42.99 & 49.02 & 46.35 & 46.44 & 46.82 \\
    \multirow{4}{*}{\rotatebox[origin=c]{90}{\parbox[c]{1cm}{\centering Proposed}}}
    &Type3 & 42.13 & 43.02 & 49.49 & 45.76 & 46.38 & 46.35 \\
    &Type4 & 43.01 & 43.25 & 48.12 & 46.05 & 45.89 & 45.05 \\
    &Type5 & 42.29 & 43.34 & 49.19 & 46.09 & 46.82 & 47.21 \\
    &Type6 & 43.08 & 43.73 & 49.98 & 46.95 & 47.20 & 47.43 \\
    &Type7 & 43.16 & 43.55 & 49.28 & 46.19 & 46.70 & 46.50 \\
    &Final SGM & \textbf{43.88} & \textbf{44.37} & \textbf{51.04} & \textbf{47.68} & \textbf{48.05} & \textbf{48.69} \\ \bottomrule
    \end{tabular}
\end{table}
\begin{figure*}[th]
    \centering
    \includegraphics[width=\linewidth]{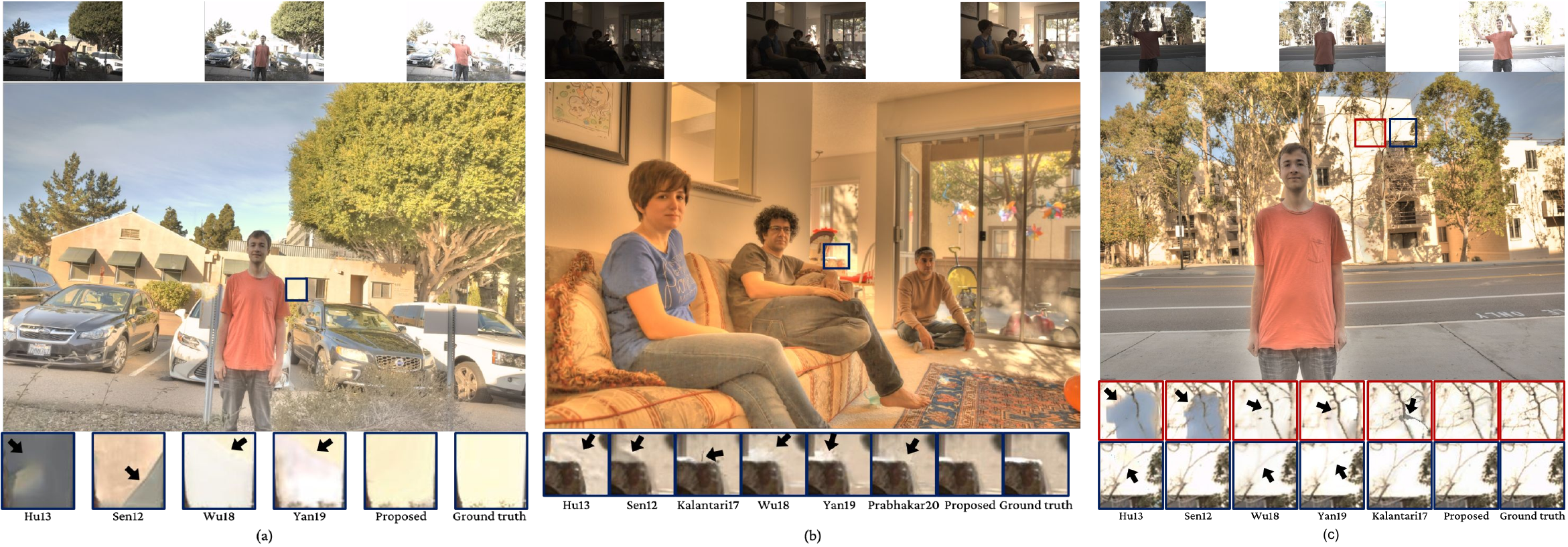}
    \caption{Qualitative comparison between state-of-the-art approaches and the proposed approach on three validation sequences from Kalantari17 dataset \cite{kalantari2017deep}.}
    \label{fig:ucsd4}
\end{figure*}
To show the contribution of each structure and connection within the cell, we perform a series of architecture experiments with the particular component removed. The following ablations are performed:
\begin{itemize}[wide,itemindent=1em,noitemsep,topsep=0pt,itemsep=0pt]
    \item \textbf{Type 1}: We remove the connection feeding $c_{n-1}$ to the remaining inputs of the output gate. The output of the cell now becomes a function of the input of the current internal state and input, with the information of previous inputs present as a nonlinear transformation within the new state. We observe that removing this connection results in a minor dip in overall performance of the network.
    \item \textbf{Type 2}: We do not concatenate $E_n$ to the other inputs of the output gate. The output is thus derived only from the new state that integrates information from the current input, and previous state. Important information about the specific frame may be lost during this computation. As such, the drop in performance after removing this connection is significant.
    \item \textbf{Type 3}: The output gate only takes $c_n$ as its input; $c_{n-1}$ and $E_n$ are not used. As expected, drop in overall performance is greater than removing any of the individual connections.
    \item \textbf{Type 4}: We remove the transform gate completely from the cell. The remaining parts are same as the final version. The drop in performance indicates that it is advantageous to have some of the local information in $c_{n-1}$ forgotten before the state update, without interference from the new input.
    \item \textbf{Type 5}: We replace self-gating with two separate layers having sigmoid and Tanh as their activations, and multiply them to get the output of the gate. We follow the LSTM-style connections in the output gate. This shows that it is harder to optimize gates having two separate components.
    \item \textbf{Type 6}: We perform LSTM-style connections to ``forget" parts of $c_{n-1}$. That is, $c_{n-1}$ is not used to compute the weightmaps to perform state update. 
    \item \textbf{Type 7}: We evaluate whether the order of processing $\{E_n, h_{n-1}\}$ and forgetting from $c_{n-1}$ impacts performance. We modify the cell structure from Type 6 ablation to perform this experiment.
\end{itemize}
\section{Evaluation and Results}
    \label{sec:eval}


\begin{table*}[ht]
        \caption{Quantitative comparison between the proposed method against nine state-of-the-art methods. The best score is highlighted by blue cell color, and the second-best score is highlighted by gray cell color. The abbrevations used are as follows: T1 - linear domain, T2 - log tonemapper, T3 - Krawczyk \textit{et al.} \cite{krawczyk2005lightness}, T4 - Reinhard \textit{et al.} \cite{reinhard2005dynamic}, and T5 - Durand \textit{et al.} \cite{durand2002fast} tonemappers.}
        \setlength{\tabcolsep}{2pt}
        \centering
        \subfloat[][Quantitative evaluation on \cite{kalantari2017deep} dataset.]{
        \label{tab:ucsd_eval}
        \begin{tabular}{@{}lccccccccccc@{}}
        \toprule
        \multicolumn{1}{c}{} & \multicolumn{5}{c}{PSNR} &  \multicolumn{5}{c}{SSIM} &  \\ \cmidrule(lr){2-6} \cmidrule(lr){7-11}
        \multicolumn{1}{c}{} & T1 & T2 & T3 & T4 & T5 & T1 & T2 & T3 & T4 & T5 & \multirow{-2}{*}{\begin{tabular}[c]{@{}c@{}}HDR-\\ VDP-2\end{tabular}} \\ \midrule
        Hu13 & 31.25 & 35.75 & 30.18 & 28.87 & 31.69 &  0.941 & 0.963 & 0.940 & 0.942 & 0.944 & 62.07 \\
        Sen12 & 38.57 & 40.94 & 27.81 & 30.33 & 31.35 &  0.971 & 0.978 & 0.955 & 0.966 & 0.969 & 64.74 \\
        Endo17 & 8.846 & 21.33 & 12.26 & 12.10 & 20.65 &  0.107 & 0.622 & 0.715 & 0.787 & 0.738 & 54.00 \\
        Eilertsen17 & 14.21 & 14.13 & 25.23 & 20.26 & 26.82 & 0.350 & 0.882 & 0.925 & 0.923 & 0.935 & 57.95 \\
        Kalantari17 & 41.27 & \cellcolor[HTML]{C0C0C0}42.74 & 34.12 & 33.70 & 32.99 & 0.981 & 0.987 & 0.980 & 0.979 & 0.974 & 66.10 \\
        Wu18 & 40.91 & 41.65 & 34.98 & 34.54 & 33.69 & 0.986 & 0.986 & 0.982 & 0.982 & 0.977 & 67.44 \\
        Prabhakar19 & 39.68 & 40.47 & 33.56 & 34.08 & 32.60 &  0.980 & 0.975 & 0.966 & 0.975 & 0.961 & 66.50 \\
        Prabhakar20 & \cellcolor[HTML]{C0C0C0}41.33 & \cellcolor[HTML]{A0A2ED}\textbf{42.82} & \cellcolor[HTML]{C0C0C0}35.18 & \cellcolor[HTML]{C0C0C0}36.94 & \cellcolor[HTML]{C0C0C0}36.29 & 0.986 & \cellcolor[HTML]{C0C0C0}0.989 & \cellcolor[HTML]{C0C0C0}0.984 & \cellcolor[HTML]{C0C0C0}0.985 & \cellcolor[HTML]{C0C0C0}0.982 & 67.15 \\
        Yan19 & 41.08 & 41.21 & 29.83 & 33.28 & 29.51 &  \cellcolor[HTML]{C0C0C0}0.989 & \cellcolor[HTML]{C0C0C0}0.989 & 0.962 & 0.978 & 0.974 & \cellcolor[HTML]{C0C0C0}67.53 \\
        Proposed    & \cellcolor[HTML]{A0A2ED}\textbf{41.68} & 42.07 & \cellcolor[HTML]{A0A2ED}\textbf{36.09} & \cellcolor[HTML]{A0A2ED}\textbf{37.85} & \cellcolor[HTML]{A0A2ED}\textbf{36.29} &  \cellcolor[HTML]{A0A2ED}\textbf{0.990} & \cellcolor[HTML]{A0A2ED}\textbf{0.990} & \cellcolor[HTML]{A0A2ED}\textbf{0.986} & \cellcolor[HTML]{A0A2ED}\textbf{0.988} & \cellcolor[HTML]{A0A2ED}\textbf{0.983} & \cellcolor[HTML]{A0A2ED}\textbf{67.59} \\ \bottomrule
        \end{tabular}}
        \subfloat[][Quantitative evaluation on \cite{prabhakar2019fast} dataset.]{
        \label{tab:iccp_eval}
        \begin{tabular}{@{}ccccccccccc@{}}
        \toprule
        \multicolumn{5}{c}{PSNR} &  \multicolumn{5}{c}{SSIM} & \\ \cmidrule(lr){1-5} \cmidrule(lr){6-10}
        T1 & T2 & T3 & T4 & T5 &  T1 & T2 & T3 & T4 & T5 & \multirow{-2}{*}{\begin{tabular}[c]{@{}c@{}}HDR-\\ VDP-2\end{tabular}} \\ \midrule
        29.47 & 32.58 & 27.61 & 26.84 & 26.20 &   0.954 & 0.949 & 0.912 & 0.919 & 0.917 & 63.50 \\
        32.93 & 33.43 & 30.22 & 29.15 & 30.91 &   0.972 & 0.964 & 0.950 & 0.948 & 0.950 & 65.47 \\
        9.760 & 8.980 & 11.18 & 13.07 & 20.60 &   0.132 & 0.641 & 0.675 & 0.763 & 0.711 & 55.76 \\
        14.19 & 15.66 & 22.47 & 22.04 & 24.97 &   0.442 & 0.869 & 0.879 & 0.897 & 0.904 & 58.74 \\
        32.50 & 35.63 & 30.08 & 28.81 & 30.45 &   0.969 & 0.961 & 0.938 & 0.943 & 0.940 & 65.40 \\
        34.40 & 38.03 & 33.35 & 31.82 & 32.66 &  0.977 & 0.971 & 0.962 & 0.957 & 0.956 & 66.59 \\
        32.74 & 36.08 & 30.66 & 29.83 & 30.54 &  0.967 & 0.959 & 0.942 & 0.935 & 0.939 & 66.10 \\
        34.98 & 38.30 & 32.99 & 31.76 & 32.53 &  0.978 & 0.970 & 0.960 & 0.952 & 0.953 & 66.25 \\
        \cellcolor[HTML]{C0C0C0}35.28 & \cellcolor[HTML]{C0C0C0}38.65 & \cellcolor[HTML]{C0C0C0}33.82 & \cellcolor[HTML]{C0C0C0}32.08 & \cellcolor[HTML]{A0A2ED}\textbf{33.31} &  \cellcolor[HTML]{C0C0C0}0.980 & \cellcolor[HTML]{C0C0C0}0.973 & \cellcolor[HTML]{C0C0C0}0.963 & \cellcolor[HTML]{C0C0C0}0.961 & \cellcolor[HTML]{C0C0C0}0.957 & \cellcolor[HTML]{C0C0C0}66.88 \\
        \cellcolor[HTML]{A0A2ED}\textbf{36.38} & \cellcolor[HTML]{A0A2ED}\textbf{39.03} & \cellcolor[HTML]{A0A2ED}\textbf{34.45} & \cellcolor[HTML]{A0A2ED}\textbf{32.43} & \cellcolor[HTML]{C0C0C0}32.85 &   \cellcolor[HTML]{A0A2ED}\textbf{0.983} & \cellcolor[HTML]{A0A2ED}\textbf{0.975} & \cellcolor[HTML]{A0A2ED}\textbf{0.967} & \cellcolor[HTML]{A0A2ED}\textbf{0.963} & \cellcolor[HTML]{A0A2ED}\textbf{0.960} & \cellcolor[HTML]{A0A2ED}\textbf{67.95} \\ \bottomrule
        \end{tabular}}
        \label{tab:eval}
    \end{table*}
\subsection{Implementation}
Our models and pipelines were implemented using Tensorflow. The network was trained on a machine with Intel core i7-8700 CPU and a NVIDIA RTX 2080ti 11GB GPU. We use the Adam optimizer \cite{kingma2014adam} with $2\times10^{-4}$ learning rate and a batch size of four to train the model for 200 epochs. The learning rate is halved after every 25 epochs. We train the model using Kalantari \textit{et al.} \cite{kalantari2017deep} and Prabhakar \textit{et al.} \cite{prabhakar2019fast} datasets for 3-image fusion. \cite{kalantari2017deep} dataset consists of 74 training and 15 validation sequences, while \cite{prabhakar2019fast} dataset consists of 466 training and 116 validation sequences. In both \cite{kalantari2017deep} and \cite{prabhakar2019fast} datasets, each sequence consists of three varying exposure images with exposure bias $\{-2,0,+2\}$ or $\{-3,0,+3\}$. Additionally, we trained on Prabhakar \textit{et al.} \cite{prabhakar2020hdrdb} dataset for validating scalability performance. Dataset given by \cite{prabhakar2020hdrdb} consists of 70 training sequences with 7 LDR images, and the sequence length (3, 5 or 7) is selected randomly during training. It also contains 14 sequences with 7 LDR images for testing. The sequence $\{I_2, I_4, I_6\}$ is used for validating 3-image performance and sequence $\{I_2, I_3, I_4, I_5, I_6\}$ is used for performing 5-image validation. Images ${I_1}$ through ${I_7}$ are all used for 7-image validation. Apart from \cite{kalantari2017deep}, \cite{prabhakar2019fast} and \cite{prabhakar2020hdrdb}, we have validated our model on publicly available datasets like Sen \textit{et al.} \cite{sen2012robust}, Cambridge \cite{karaduzovic2017multi} and Tursun \textit{et al.} \cite{tursun2016objective}.
\subsection{Quantitative results}
In Table \ref{tab:eval}, we compare the proposed method trained for 3 images on \cite{kalantari2017deep} and \cite{prabhakar2019fast} datasets against nine other HDR deghosting approaches. The approaches against which we compare the proposed approach are: \begin{enumerate*}[label=(\roman*)]
      \item Hu13 - Hu \textit{et al.} \cite{hu2013hdr},
      \item Sen12 - Sen \textit{et al.} \cite{sen2012robust},
      \item Endo17 - Endo \textit{et al.} \cite{endo2017deep},
      \item Eilertsen17 - Eilertsen \textit{et al.} \cite{eilertsen2017hdr},
      \item Kalantari17 - Kalantari \textit{et al.} \cite{kalantari2017deep},
      \item Wu18 - Wu \textit{et al.} \cite{wu2018deep},
      \item Prabhakar19 - Prabhakar \textit{et al.} \cite{prabhakar2019fast},
      \item Prabhakar20 - Prabhakar \textit{et al.} \cite{prabhakar2020}
      \item Yan19 - Yan \textit{et al.} \cite{yan2019attention}.
    \end{enumerate*}
We evaluate the performance of different methods using three popular metrics: HDR-VDP-2 \cite{mantiuk2011hdr}, SSIM \cite{wang2004image} and PSNR. HDR-VDP-2 is a full reference HDR image quality assessment metric for evaluating prediction quality in all luminance ranges.

To get an overall view of quality, we compute SSIM and PSNR using four different tonemapping functions, in addition to comparison in the linear domain. The tonemapping functions used are: Log tonemapper (Eqn. \ref{eqn:log_tm}), Krawczyk \textit{et al.} \cite{krawczyk2005lightness}, Reinhard \textit{et al.} \cite{reinhard2002photographic} and Durand \textit{et al.} \cite{durand2002fast}. In total, we compare against nine existing methods on eleven different metrics. Table \ref{tab:eval}a shows quantitative comparison on Kalantari17 dataset with 15 test sequences. Our proposed method outperforms all other existing methods on ten out of eleven metrics. Similarly, in Table \ref{tab:eval}b, we present a quantitative comparison on Prabhakar19 dataset with 116 test images. We have trained all existing methods and our method on \cite{prabhakar2019fast} for a fair comparison. The proposed approach outperforms all approaches in every metric except for PSNR on the \cite{durand2002fast} tonemapping function.
\begin{table}[t]
    \setlength{\tabcolsep}{3pt}
    \centering
    \caption{Comparison of the proposed method against existing methods to fuse variable number of images from \cite{prabhakar2020hdrdb} dataset.}
    \label{tab:ablation_arbitrary_length}
    \begin{tabular}{@{}llcccc|cccc@{}}
    \toprule
    \multicolumn{2}{c}{Metrics ($\rightarrow$)} & \multicolumn{4}{c|}{PSNR-L} & \multicolumn{4}{c}{PSNR-T} \\ \midrule
    \multicolumn{2}{c}{\begin{tabular}[c]{@{}c@{}}Images ($\rightarrow$)\\ / Methods ($\downarrow$)\end{tabular}} & 3 & 5 & 7 & Mean & 3 & 5 & 7 & Mean \\ \midrule
    \multicolumn{2}{l}{Sen12} & 40.61 & 41.71 & 47.08 & 43.13 & 46.26 & 47.51 & 47.27 & 47.01 \\
    \multicolumn{2}{l}{Hu13} & 35.71 & 36.38 & 39.31 & 37.13 & 41.85 & 42.65 & 40.81 & 41.77 \\
    \multicolumn{2}{l}{Prabhakar19} & 39.53 & 39.47 & 45.43 & 41.47 & 39.79 & 40.02 & 40.10 & 39.97 \\ \midrule
    \multirow{4}{*}{\rotatebox[origin=c]{90}{\parbox[c]{1cm}{\centering Proposed}}}&Bi-Vanilla & 43.26 & 43.88 & 50.07 & 45.74 & 46.15 & 46.52 & 47.16 & 46.61 \\
    &Bi-LSTM & 43.10 & 43.90 & 50.30 & 45.77 & 47.02 & 47.63 & 48.28 & 47.64 \\
    &Bi-GRU & 42.46 & 43.02 & 49.23 & 44.90 & 45.69 & 46.01 & 46.45 & 46.05 \\
    &Bi-SGM & \textbf{43.88} & \textbf{44.37} & \textbf{51.04} & \textbf{46.43} & \textbf{47.68} & \textbf{48.05} & \textbf{48.69} & \textbf{48.14} \\ \bottomrule
    \end{tabular}
\end{table}

In Table \ref{tab:ablation_arbitrary_length}, we present quantitative comparison against existing scalable HDR deghosting methods on Prabhakar \textit{et al.} dataset \cite{prabhakar2020hdrdb} with sequences of length 3,5 and 7. We also compare against different recurrent cells to demonstrate the proposed SGM cell's effectiveness.
\subsection{Qualitative results}
Among the classical non-deep HDR deghosting methods, Hu13 \cite{hu2013hdr} and Sen12 \cite{sen2012robust} are popular patch-based optimization approaches\footnote{Please refer to Supplementary material for more results.}. As both Hu13 and Sen12 methods synthesize the result with reference images, these methods' success depends on carefully choosing a reference image with the least amount of saturation. In the presence of heavily saturated regions in the reference image, they tend to introduce artifacts. As shown in Fig. \ref{fig:seq7_1}, the result by Hu13 method has structural artifacts in heavily saturated regions of the reference image. In the same region, the Sen12 method generates results without any texture information. Similarly, in \ref{fig:ucsd4}c, we present a qualitative comparison between different state-of-the-art methods and proposed method on a challenging sequence from the Kalantari17 dataset. The results in Fig. \ref{fig:ucsd4} and \ref{fig:seq7_2} show the artifacts introduced by both Hu13 and Sen12 methods on various sequences with arbitrary length sequences.  

Fig. \ref{fig:ucsd4}c shows a qualitative comparison between Kalantari17 and the proposed method on a difficult example from the Kalantari17 dataset. The Kalantari17 method has residual optical flow warping artifacts present in the result shown in the zoomed red box. Similarly, in Fig. \ref{fig:ucsd4}b, their result has warping artifact along the boundaries. In another example from the Prabhakar19 dataset shown in Fig. \ref{fig:iccp2}c, Kalantari17 method fails to correct warping artifacts in the dynamic regions. In Fig. \ref{fig:iccp2}a, Kalantari17 method's result has artifacts even in static regions.

As shown in Fig. \ref{fig:ucsd4}c, \ref{fig:ucsd4} and \ref{fig:iccp2}, Wu18 and Yan19 methods suffer from artifacts in reference saturated dynamic regions. In such regions, Wu18 and Yan19 methods fail to reconstruct complete texture from other LDR input images. In \ref{fig:ucsd4}c, both Wu18 and Yan19 generate results with missing branches. Also, in Fig. \ref{fig:ucsd4}, both methods introduce artifacts in regions affection by saturation and motion. Similarly, in Fig. \ref{fig:iccp2}a and Fig. \ref{fig:iccp2}b, Wu18 and Yan19 results fail to faithfully reconstruct window bars; instead, they smoothen out the details. 
\begin{figure}[t]
    \centering
    \includegraphics[width=\linewidth]{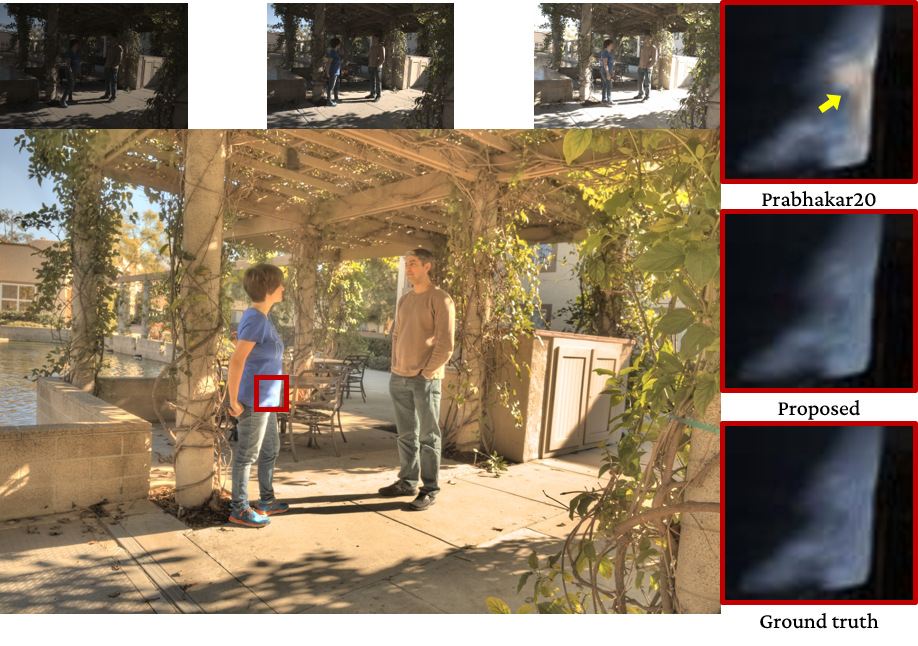}
    \caption{A qualitative example from Kalantari17 dataset \cite{kalantari2017deep}. The red zoomed box highlight the artifact introduced by the Prabhakar20 method on saturated dynamic regions. Comparatively, the proposed method can generate plausible textures, even in saturated areas. Image best viewed electronically.}
    \label{fig:ucsd1}
\end{figure}

Prabhakar20 method generates output with artifacts as shown in Fig. \ref{fig:ucsd1}. The zoomed highlighted region is affected by motion and saturation. As the Prabhakar20 method's input is optical flow corrected, their approach hallucinates incorrect textures in the saturated areas. In Fig. \ref{fig:ucsd4}b, a similar artifact is noticed in the saturated and moving region. In Fig. \ref{fig:iccp2}c, both Prabhakar20 and Kalantari17 methods result in artifacts due to optical flow alignment. In Fig. \ref{fig:iccp2}a and Fig. \ref{fig:iccp2}b, Prabhakar20 method has residual ghosting artifact left in the result. 
\begin{figure*}[th]
    \centering
    \includegraphics[width=\linewidth]{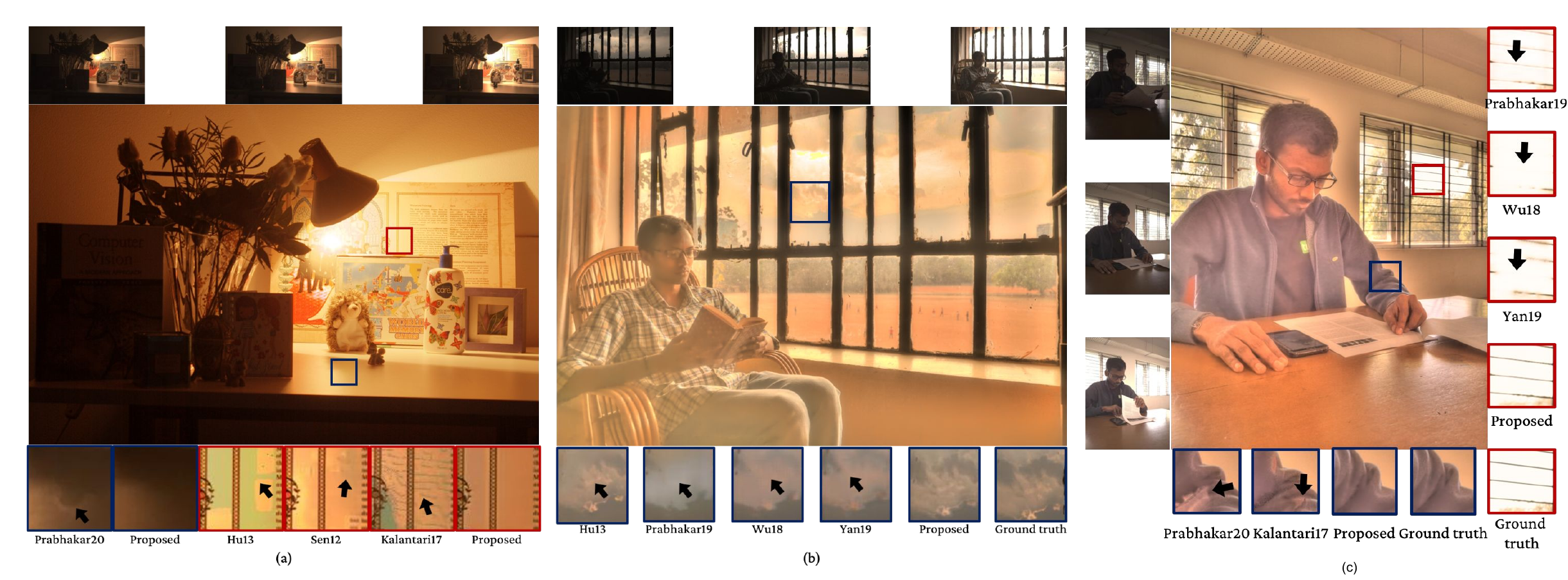}
    \caption{Qualitative comparison on sequences from (a) Cambridge dataset \cite{karaduzovic2017multi}, (b,c) Prabhakar19 dataset \cite{prabhakar2019fast}. Image best viewed electronically.}
    \label{fig:iccp2}
\end{figure*}

Prabhakar19 is the only scalable deep learning-based method. In Fig. \ref{fig:seq7_1}, we show an example with seven input images. Prabhakar19 method's output has color distortion in the region marked with a blue box. In Fig. \ref{fig:iccp2}c, the Prabhakar19 method's output has an over-smoothing artifact in the saturated reference region. Similarly, in Fig. \ref{fig:seq7_2}a the output of Prabhakar19 method contains inaccurate texture details. In Fig. \ref{fig:iccp2}b, Prabhakar19 method over-smoothens the cloud region, thus losing the necessary details in them.

In comparison, the proposed method generates results with accurate texture and vivid color details for variable-length sequences without re-training. The proposed method's result is void of ghosting artifacts and can generate plausible textures in regions affected by motion and saturation (Fig. \ref{fig:ucsd1} and \ref{fig:ucsd4}). It should be noted that, as the existing state-of-the-art deep learning methods except Prabhakar19, were trained on three input images, we tested those methods only on sequences with three input images. 

In Fig. \ref{fig:rebuttal_fig6}, we present an example for adding more LDR images in a dynamic scene. As seen from the results, adding more images leads to increase in the dynamic range. 
\begin{figure}
    \centering
    \includegraphics[width=\linewidth]{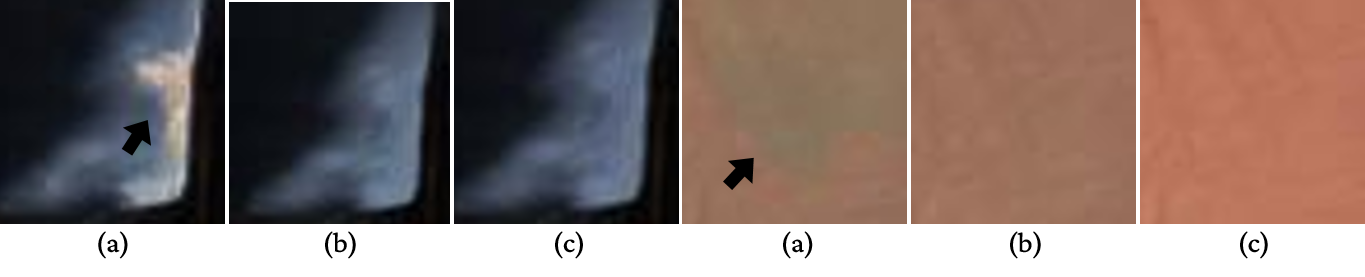}
    \caption{Qualitative comparison between (a) unidirectional and (b) bidirectional SGM cells. As highlighted by the black arrow, the output by unidirectional cell has artifacts. The ground truth is shown in (c).}
    \label{fig:uni_bi_ablation}
\end{figure}


\begin{figure*}[t]
        \centering
        \includegraphics[width=\linewidth]{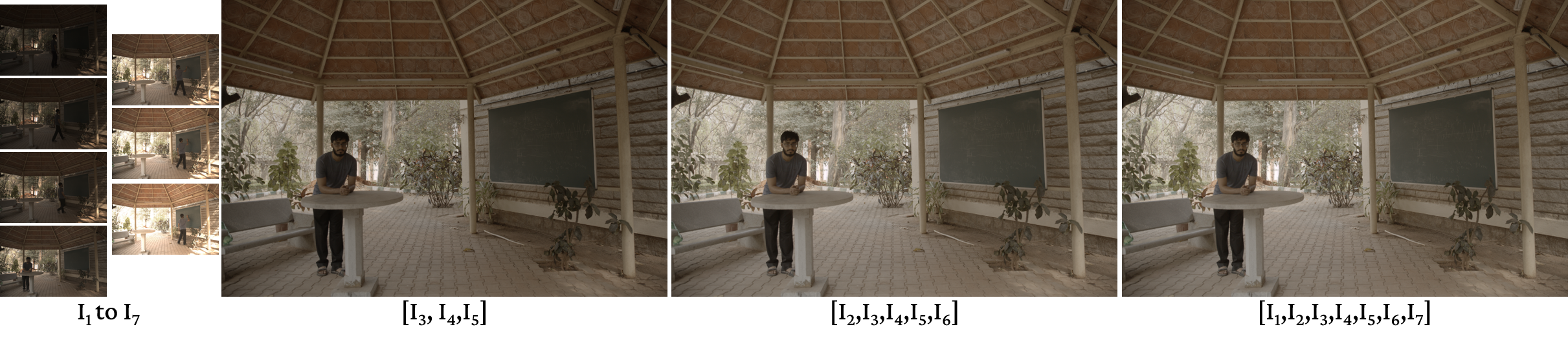}
        \caption{An qualitative example to highlight effect of adding more LDRs to the network. Left column: seven input varying exposure images with foreground motion. Second column image: output by proposed method by fusing middle three images ($I_3, I_4, I_5$). Third column image: by fusing middle five images $I_2$ to $I_6$. Last image: by fusing all seven images. }
        \label{fig:rebuttal_fig6}
    \end{figure*}
    \label{sec:discussion}

\subsection{Ablations}
\textit{Uni vs Bi-directional SGM}: We run various experiments to show the effectiveness of our approach in a variety of baseline settings. To validate the importance of bi-directional setting, we train a model using only one SGM cell in one direction. The input to decoder consists of final time step feature maps. The results are provided in Table \ref{tab:ablations}. Also, we observe that using optical flow provides slight boost across all metrics. We observe that bi-directional SGM cell offers better results compared to uni-directional cell in all three comparison metrics. The qualitative comparison in Fig. \ref{fig:uni_bi_ablation} shows that the output by uni-directional model has visible artifacts in saturated regions, which is corrected in bi-directional model.

\begin{figure*}[h]
    \centering
    \includegraphics[width=\linewidth]{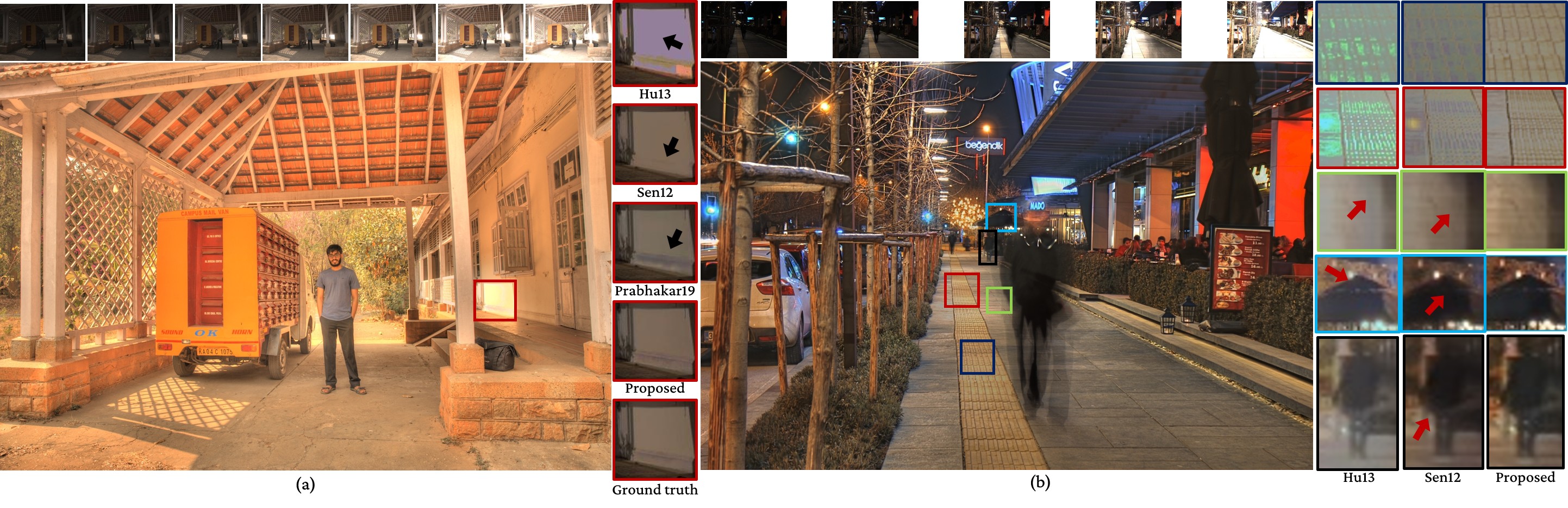}
    \caption{A qualitative example from (a) Prabhakar19 dataset \cite{prabhakar2019fast}, and (b) METU dataset \cite{tursun2015state}.}
    \label{fig:seq7_2}
\end{figure*}

\textit{Scalability}: To verify the effectiveness of our model to fuse arbitrary number of images without retraining, we have tested a model trained with UCSD dataset \cite{kalantari2017deep} on \cite{prabhakar2020hdrdb} dataset without retraining for different input image lengths. We achieved a average PSNR-L of 43.64dB, while the only other learnable scalable architecture \cite{prabhakar2019fast} achieves 38.02dB. Also, we show qualitative comparisons for HDR Deghosting on the METU dataset consisting of 9-image sequences without re-training (Fig. \ref{fig:seq7_2}). Additionally, we also show qualitative results for the Multi-Exposure fusion problem using our method, where the models have been trained on sequence lengths ranging from 3 images to 9 images and tested on examples having 30 images in a sequence (Fig. 9 in supplementary file). 
\begin{table}[t]
    \centering
    \caption{Quantitative comparison between several baseline ablation experiments on Kalantari17 dataset.}
    \label{tab:ablations}
    \begin{tabular}{@{}cccccc@{}}
    \toprule
    \begin{tabular}[c]{@{}c@{}}Uni/Bi\\ directional\end{tabular} & \begin{tabular}[c]{@{}c@{}}Optical\\ flow\end{tabular} & PSNR-L & PSNR-T & \begin{tabular}[c]{@{}c@{}}HDR-\\ VDP-2\end{tabular} \\ \midrule
    \multirow{2}{*}{$\rightarrow$}      & \xmark & 41.16 & 41.71 & 67.29 \\ 
      & \cmark & 41.64 & 41.84 & 67.35 \\ \midrule
    \multirow{2}{*}{$\rightleftarrows$} & \xmark & 41.26 & 41.93 & 67.33 \\
      & \cmark & \textbf{41.68} & \textbf{42.07} & \textbf{67.59} \\ \bottomrule
    \end{tabular}
\end{table} 

\textit{Image order}: In order to understand the importance of image order, we train a model after randomly shuffling the low, medium and high exposure images in the \cite{kalantari2017deep} dataset. In this experiment, we have considered all permutations of the three sequences, including those where the reference image was displaced from the central spot. The medium exposure image is used as reference, however its location in the sequence is random. For validation, the same strategy is used. The setup with optical flow corrected images gives a PSNR-L of 41.53 dB, which is comparable to existing state-of-the-art methods (Table \ref{tab:various_ablations}). This shows that our model can provide high quality HDR images even if the input LDR images are not sorted by exposure time. We have also trained a model with 5 images in random order which achieved a PSNR-L of 43.62dB versus 43.75dB without shuffling. 
     

\textit{SGM cell}: We also conduct experiments regarding the structure of the proposed SGM Cell (Table \ref{tab:various_ablations}). They are not included as separate Type Ablations, since they are minimal changes on top of the final version of the cell. In the first experiment, we replace the swish activation at the output gate with the Tanh activation (change in equations 19 and 24), and observe a drop of 0.94 dB in PSNR-L. In another ablation experiment, we remove the Swish activations and use the outputs of the gates in equations \eqref{eqn:eqn15}, \eqref{eqn:eqn16} and \eqref{eqn:eqn19} (and corresponding reverse equations) directly without applying any nonlinearity. This experiment resulted in a drop of 0.63 dB in PSNR-L. 

\begin{table}[t]
\caption{An ablation study on the importance of SGM cell components.}
\label{tab:various_ablations}
\centering
\begin{tabular}{@{}lc@{}}
\toprule
\multicolumn{1}{c}{Ablations} & PSNR-L (dB)\\ \midrule
Shuffled input sequence & 41.53 \\
Tanh at the output gate & 40.74 \\
Without Sigmoid in Eqn. \eqref{eqn:eqn16}, \eqref{eqn:eqn17} and \eqref{eqn:eqn19} & 41.05 \\
Proposed final architecture & \textbf{41.68} \\ \bottomrule
\end{tabular}
\end{table}
\subsection{Running times} 
In Table \ref{tab:running_time}, we present the running time comparison between existing methods and the proposed method for variable length sequences. Our approach can fuse a sequence with three images of resolution 1000$\times$1500 in 0.03 seconds on a NVIDIA Quadro RTX 6000 GPU and a Intel Core i7 3.00 GHz CPU. Also, our method is atleast 17$\times$ faster than other scalable HDR deghosting approaches. 
\begin{table}[t]
\centering
\caption{Running time (in seconds) and number of parameters comparison between existing methods and proposed method.}
\label{tab:running_time}
\begin{tabular}{@{}llccc|cc@{}}
\toprule
\multicolumn{2}{c}{} & \multicolumn{3}{c|}{Number of Images} & \multicolumn{2}{c}{Parameters (Millions)} \\ \midrule
\multicolumn{2}{c}{\begin{tabular}[c]{@{}c@{}}Architecture\end{tabular}} & 3 & 5 & 7 & Network & Cell \\ \midrule
\multicolumn{2}{l}{Sen12}     & 146.0  & 496  & 1187  & - & -\\
\multicolumn{2}{l}{Hu13}      & 264.2  & 550   & 798  & - & -\\
\multicolumn{2}{l}{Kalantari17} & 50.02 & -  &  -   & 0.38 & -\\
\multicolumn{2}{l}{Wu18}      & 6.800 & -     & -    & 16.61 & -\\
\multicolumn{2}{l}{Yan19}     & 0.096 & - & - & 1.44 & -\\
\multicolumn{2}{l}{Prabhakar19} & 0.663 & 1.274 & 1.873 & 12.21 & -\\ \midrule
\multirow{3}{*}{\rotatebox[origin=c]{90}{\parbox[c]{1cm}{\centering Proposed}}}&Bi-LSTM   & 0.042 & 0.058 & 0.075 & 1.30 & 0.33\\
& Bi-GRU    & 0.032 & 0.049 & 0.059 & 1.12 & 0.22\\
& Bi-SGM & 0.037 & 0.052 & 0.072 & 1.19 & 0.26\\ \bottomrule
\end{tabular}
\end{table}

\section{Conclusion}
    \label{sec:conc}
    We propose a efficient scalable HDR deghosting method based on recurrent neural networks. In our approach, we introduce a novel Self-Gated Memory recurrent cell that can control the information flow by gating the output with a function of itself. The SGM cell offers high accuracy than standard GRU and has less parameters than LSTM cell. We utilize SGM cell in bi-directional setting to achieve better performance and visually pleasing results. The major advantage of our proposed method compared to many existing deep learning based HDR deghosting methods, is that our approach can fuse arbitrary length sequence without a need for re-training. Further, we demonstrate the superiority of our method over existing state-of-the-art methods on three publicly available datasets. 

%



\section*{Acknowledgment}
This work was supported by a project grant from MeitY (No.4(16)/2019-ITEA) and Uchhatar Avishkar Yojana (UAY) project (IISC\_10), MHRD, Govt. of India.





\bibliographystyle{IEEEtran}
\bibliography{main.bib}

%








\end{document}